%% file: main.tex
\documentclass[10pt, conference, compsocconf]{IEEEtran}
\IEEEoverridecommandlockouts
\usepackage{cite}
\usepackage{amsmath,amssymb,amsfonts}
\usepackage{graphicx}
\usepackage{textcomp}
\usepackage{xcolor}
\usepackage{subfigure}
\usepackage{algorithm}
\usepackage{algpseudocode}
\usepackage{xspace}
\usepackage{booktabs}
\usepackage{array}
\newcolumntype{C}[1]{>{\centering\arraybackslash}p{#1}}
\usepackage[draft]{hyperref}

\def\BibTeX{{\rm B\kern-.05em{\sc i\kern-.025em b}\kern-.08em
    T\kern-.1667em\lower.7ex\hbox{E}\kern-.125emX}}

\newcommand{\sysname}{MobileARLoc\xspace}

\begin{document}

\title{MobileARLoc: On-device Robust Absolute Localisation for Pervasive Markerless Mobile AR
}

\author{Changkun Liu\textsuperscript{1*}, Yukun Zhao\textsuperscript{1*}, Tristan Braud\textsuperscript{1,2}\\
\IEEEauthorblockA{\textsuperscript{1}\textit{Department of Computer Science and Engineering, The Hong Kong University of Science and Technology}, Hong Kong\\
\textsuperscript{2}\textit{Division of Integrative Systems and Design, The Hong Kong University of Science and Technology}, Hong Kong\\
\{cliudg,yzhaoeg\}@connect.ust.hk, braudt@ust.hk
}
\thanks{\textsuperscript{*}Both authors contributed equally to this research.}
}%

\maketitle

\begin{abstract}
 
Recent years have seen significant improvement in absolute camera pose estimation, paving the way for pervasive markerless Augmented Reality (AR). However, accurate absolute pose estimation techniques are computation- and storage-heavy, requiring computation offloading. As such, AR systems rely on visual-inertial odometry (VIO) to track the device's relative pose between requests to the server. However, VIO suffers from drift, requiring frequent absolute repositioning. This paper introduces \sysname, a new framework for on-device large-scale markerless mobile AR that combines an absolute pose regressor (APR) with a local VIO tracking system. Absolute pose regressors (APRs) provide fast on-device pose estimation at the cost of reduced accuracy. To address APR accuracy and reduce VIO drift, \sysname creates a feedback loop where VIO pose estimations refine the APR predictions. The VIO system identifies reliable predictions of APR, which are then used to compensate for the VIO drift. We comprehensively evaluate \sysname through dataset simulations. \sysname halves the error compared to the underlying APR and achieves fast (80\,ms) on-device inference speed.
\end{abstract}

\input{Sections/1Introduction}
\input{Sections/2RelatedWork}

\input{Sections/3Method}
\input{Sections/4Experiment}

\input{Sections/6Conclusion}

\bibliographystyle{IEEEtran} 
\bibliography{IEEEabrv,IEEEexample}

\end{document}

%% file: Sections/1Introduction.tex
\section{Introduction}
\label{sec:intro}

Visual localization systems utilize the visual data captured by a device's camera to determine its 6 degrees of freedom (6DOF) absolute pose (translation and rotation) within established world coordinates for a known scene. Accurate visual positioning is essential for augmented reality (AR) applications where content is anchored into the physical world without markers, paving the way to pervasive AR. 

Highly accurate localisation systems typically rely on 3D structure-based methods~\cite{dusmanu2019d2,sarlin2019coarse,taira2018inloc,noh2017large} that detect and match visual features in images against a 3D model of the environment. However, these methods are often demanding in terms of computation and storage~\cite{moreau2023imposing}. 
Scaling up such methods to larger environments thus requires offloading computations to distant servers, adding latency and  jitter~\cite{braud2020multipath}, and raising significant privacy concerns~\cite{fernandez2022implementing}.
These constraints have led the industry to adopt
on-device localisation with strict access control on the camera frames in home environments\footnote{\url{https://developer.oculus.com/blog/mixed-reality-with-passthrough/}}.
However, the computational cost of structure-based methods prevents their application on-device at a larger scale.
Absolute pose regressors (APRs) are end-to-end machine learning models that estimate the device pose using a single monocular image. They provide fast on-device inference with minimal storage, even in large environments and over multiple scenes~\cite{shavit2021learning}. However, their low accuracy and robustness have prevented their application to large-scale mobile AR~\cite{sattler2019understanding}. 
Although absolute localisation is a significant challenge, 
most markerless AR applications track the relative pose through visual-inertial odometry (VIO).
VIO systems calculate the displacement between camera frames using visual and inertial data from sensors. 
As such, they tend to display high accuracy in the short term, but they drift over time~\cite{scargill2022here}. 
APR and VIO present antagonistic features: APR poses are noisy yet drift-free, while VIO is very accurate with errors building up over time~\cite{brahmbhatt2018geometry}. We believe that VIO's high accuracy could thus improve APR's imprecision, while the APR's predictions could address VIO drift.

This paper introduces \sysname, an on-device visual localisation framework for large-scale mobile AR that combines the complementary properties of VIO and APR. 
Accurate APR predictions should be consistent with the relative odometry estimates obtained from VIO. 
Otherwise, the APR prediction should be considered unreliable.
\sysname aligns the APR and VIO coordinate systems to identify reliable APR poses and refine unreliable predictions. 
When several consecutive APR poses are consistent with the VIO relative output, they are considered accurate.
\sysname calculates the average of reliable absolute predictions as the \textit{reference pose} and the rigid transformation between this reference pose and the corresponding VIO poses to align the coordinate systems.
Following the alignment stage, \sysname enters the \textit{pose optimization stage}. Each APR pose is compared to the corresponding VIO pose.
The APR prediction is output directly if reliable. Otherwise, \sysname outputs the VIO pose converted into absolute world coordinates using rigid transformation. We introduce a new similarity metric to detect VIO drift. When drift is detected, \sysname reenters the alignment stage to select new reliable poses and calculate rigid transformation. 

We summarize our main contributions as follows:
\begin{enumerate}
\item   We \textbf{design an APR-agnostic framework}, \sysname, for real-time on-device pose estimation. \sysname leverages VIO data 
to select reliable APR predictions, refine unreliable predictions and compensate drift. 

\item We \textbf{implement and evaluate \sysname} over two popular APR models, PoseNet (PN)~\cite{kendall2015posenet} and MS-Transformer (MS-T)~\cite{shavit2021learning}. 
\sysname improves MS-T's accuracy by up to 47\% in translation and 66\% in rotation over the average of the three outdoor scenes. 
\item We integrate \sysname into a \textbf{real-life mobile AR application} and evaluate its performance.  

\end{enumerate}

%% file: Sections/2RelatedWork.tex
\section{Related work}
\subsection{Absolute Pose Regression}
Absolute Pose Regressors train deep neural networks to regress the 6-DOF camera pose of a query image. The first APR is PoseNet (PN)~\cite{kendall2015posenet}. Since then, there have been several improvements to APR, mainly related to the backbone architecture and loss functions~\cite{kendall2017geometric, chen2022dfnet,brahmbhatt2018geometry}.  MS-Transformer (MS-T)~\cite{shavit2021learning} extends the single-scene paradigm of APR for learning multiple scenes. MapNet+~\cite{brahmbhatt2018geometry} and DFNet$_{dm}$~\cite{chen2022dfnet} finetune pre-trained network on unlabeled test data to improve the accuracy.
However, in-test-time finetuning neural networks is time-consuming and unlabeled test-set data is difficult to obtain in advance in real applications. Among these works, MapNet~\cite{brahmbhatt2018geometry} aims to 
minimize the loss of the per-image absolute pose and the loss of the relative pose between image pairs. 
However, formulating relative pose constraints as loss terms during training shows limited accuracy improvement and has been surpassed by state-of-the-art (SOTA) APRs like MS-T~\cite{shavit2021learning}. Our framework improves accuracy at test time by using the relative pose independently from the training strategy. Compared to most modern approaches, \sysname does not rely on additional unlabeled test data.

\subsection{Uncertainty estimation and Pose Optimization}
APRs suffer from limited generalizability~\cite{sattler2019understanding}. Uncertainty-aware APRs aim to infer which images will likely result in accurate pose estimation and identify the outliers.
Several prior works have explored uncertainty estimation during  APR training. Bayesian PoseNet~\cite{kendall2016modelling} and AD-PoseNet~\cite{huang2019prior} model the uncertainty by measuring the variance of several inferences of the same input data. 
CoordiNet~\cite {moreau2022coordinet} models heteroscedastic uncertainty during training.  Deng \textit{et al.}~\cite{bui20206d} represent uncertainty by predicting a mixture of multiple unimodal distributions. 
Although these uncertainty-aware APRs provide both pose predictions and uncertainty estimates, the accuracy of predictions is much lower than other APRs~\cite{kendall2017geometric,chen2022dfnet,shavit2021learning}. Moreover, existing uncertainty estimation methods can be time-consuming~\cite{kendall2016modelling} and lack extensibility due to the need for specific loss functions and training schemes~\cite{bui20206d,10160466,huang2019prior,moreau2022coordinet}.


Our framework enables greater flexibility compared to existing uncertainty-aware methods by being APR-agnostic, enabling the integration of most mainstream APRs. 
Our method performs a rigid transformation of the VIO's pose to optimize unreliable APR poses directly, without iterative optimization.  It improves the accuracy of APRs with minimal overhead, enabling reliable APR usage in mobile AR applications.

%% file: Sections/3Method.tex
\section{Proposed Approach}
\label{sec:method}
\subsection{Definition}
 Given a query image $I_{i}$ in a known scene,  APR $\mathcal{R}$ outputs global translation $\mathbf{\hat{x}}_i$ and rotation $\mathbf{\hat{q}}_i$ in an established world coordinate system for the scene, so that $\mathcal{R}({I}_i) = \hat{p}_i =  <\mathbf{\hat{x}_i},\mathbf{\hat{q}}_i>$ is the estimated camera pose for $I_{i}$. The Ground Truth (GT) of $I_{i}$ in the world coordinate system is $p_i = <\mathbf{x}_i,\mathbf{q}_i>$. The camera pose of $I_{i}$ in VIO coordinate system is noted $p_i^{vio} = <\mathbf{x}_i^{vio},\mathbf{q}_i^{vio}>$.    The relative translation between two consecutive images $I_{i}$ and $I_{i+1}$ is characterized by 
 \begin{equation}
     \hat{\Delta}_{trans} (i+1,i) = ||\mathbf{\hat{x}}_{i+1}-\mathbf{\hat{x}}_i||_2,
 \end{equation}
 \begin{equation}
     \Delta_{trans}^{vio}(i+1,i) = ||\mathbf{x}^{vio}_{i+1}-\mathbf{x}^{vio}_i||_2,
 \end{equation}
 \begin{equation}
     \Delta_{trans}(i+1,i) = ||\mathbf{x}_{i+1}-\mathbf{x}_i||_2.
 \end{equation}
Similarly, we get relative rotation between $I_{i}$ and $I_{i+1}$ in degree, $\mathbf{q}^{-1}$ denotes the conjugate of $\mathbf{q}$, and we assume all quaternions are normalized: $\hat{\Delta}_{rot}(i+1,i) = 2\arccos{|\mathbf{\hat{q}}_{i+1}^{-1}\mathbf{\hat{q}}_{i}|}\frac{180}{\pi}$, $\Delta_{rot}^{vio}(i+1,i) = 2\arccos{|\mathbf{q}_{i+1}^{vio-1}\mathbf{q}_{i}^{vio}|}\frac{180}{\pi}$ and $\Delta_{rot}(i+1,i) = 2\arccos{|\mathbf{q}_{i+1}^{-1}\mathbf{q}_{i}|}\frac{180}{\pi}$.

 $\hat{u}_{i,i+1} = <\hat{\Delta}_{trans}(i,i+1), \hat{\Delta}_{rot}(i,i+1)>$ is the odometry of $I_{i}$ and $I_{i+1}$ from predicted poses of APR. $u_{i,i+1}^{vio}= <\Delta_{trans}^{vio}(i,i+1), \Delta_{rot}^{vio}(i,i+1)>$ is the odometry of $I_{i}$ and $I_{i+1}$ from the VIO system. Similarly, $u_{i,i+1} = <\Delta_{trans}(i,i+1), \Delta_{rot}(i,i+1)>$  is the GT odometry of $I_{i}$ and $I_{i+1}$. We then define the Relative Position Error (RPE) and the Relative Orientation Error (ROE) for the VIO and the APR as follows:

\begin{equation}
    \text{RPE}^{<vio,GT>}_{i,i+1}  = |\Delta_{trans}(i+1,i) - \Delta_{trans}^{vio}(i+1,i)|
\label{eq:rpe}
\end{equation}

\begin{equation}
     \text{ROE}^{<vio,GT>}_{i,i+1} = |\Delta_{rot}(i+1,i) - \Delta_{rot}^{vio}(i+1,i)|
\label{eq:roe}
\end{equation}

\begin{equation}
    \text{RPE}^{<apr,vio>}_{i,i+1}  = |\hat{\Delta}_{trans}(i+1,i) - \Delta_{trans}^{vio}(i+1,i)|
\end{equation}

\begin{equation}
     \text{ROE}^{<apr,vio>}_{i,i+1} = |\hat{\Delta}_{rot}(i+1,i) - \Delta_{rot}^{vio}(i+1,i)|
\label{eq:roe_apr_vio}
\end{equation}

\subsection{Detecting reliable pose estimations using VIO}
\label{sec:detect}
Modern VIO systems have low drift at a small temporal scale, and $u_{i,i+1}^{vio}$ tends to be very close to the GT odometry, $u_{i,i+1}$. We can assume $\text{RPE}^{<vio,GT>}_{i,i+1}$ and $\text{ROE}^{<vio,GT>}_{i,i+1}$ are almost 0.
We model the uncertainty of the APR output, $\hat{p}_{i+1}$, with $u_{i,i+1}^{vio}$, taking advantage of this property. If the $\text{RPE}^{<apr,vio>}_{i,i+1}$ and $\text{ROE}^{<apr,vio>}_{i,i+1}$ of multiple consecutive images are very small, we consider these predictions are accurate. 

We define a distance threshold $d_{th}$ for $\text{RPE}^{<apr,vio>}$  and an orientation threshold $o_{th}$ for $\text{ROE}^{<apr,vio>}$.
An estimated APR pose is considered accurate if the error close to GT within  $\frac{d_{th}}{2}$ and $\frac{o_{th}}{2}$. Given two consecutive query images $I_{i}$, $I_{i+1}$,
\begin{enumerate}
    \item Estimated poses of $I_{i}$ and $I_{i+1}$ are accurate, then $\text{RPE}^{<apr,vio>}_{i,i+1}$  and $\text{ROE}^{<apr,vio>}_{i,i+1}$ are lower than $d_{th}$ and $o_{th}$, respectively.
    \item One of the estimated pose for $I_{i}$ and $I_{i+1}$ is not accurate, then either $\text{RPE}^{<apr,vio>}_{i,i+1}$ should be larger than $d_{th}$ or $\text{ROE}^{<apr,vio>}_{i,i+1}$ should be larger than $o_{th}$.
    \item  Both estimated poses of $I_{i}$ and $I_{i+1}$ are inaccurate. However, $\text{RPE}^{<apr,vio>}_{i,i+1}$  remains lower than $d_{th}$ and $\text{ROE}^{<apr,vio>}_{i,i+1}$ lower than $o_{th}$.
    \item Both estimated poses of $I_{i}$ and $I_{i+1}$ are inaccurate, and either $\text{RPE}^{<apr,vio>}_{i,i+1}$ is larger than $d_{th}$ or $\text{ROE}^{<apr,vio>}_{i,i+1}$ is larger than $o_{th}$.
\end{enumerate}

When either $\text{RPE}^{<apr,vio>}_{i,i+1}$ or $\text{ROE}^{<apr,vio>}_{i,i+1}$ is larger than its respective threshold (case (2) and (4)), the pose is flagged as inaccurate and can thus be filtered out. Similarly, in case (1), the two poses are identified as accurate.
In case (3), two inaccurate poses are identified as accurate. Our method uses a probabilistic approach to reducing such false positives. APR error tends to be random with a large variance. As such, two consecutive images presenting a large APR error while being close to each other in the same direction as the VIO is a rare occurrence. By comparing more pairs of images, we further reduce the probability of false positive, filtering out the most unreliable predictions.
We then obtain the rigid transformation between the VIO coordinate system and the world coordinate system by using the reliable predicted poses and VIO poses. To ensure the rigid transform relationship's reliability, we calculate the average pose of selected predicted poses as \textit{reference pose}.  The rotation and translation of the coordinate system of VIO and the world coordinate system change over time due to the VIO drift. Therefore, we only need to update the reliable poses occasionally and optimize the predicted pose by calculating the new rotation and translation.

\begin{figure}[!t]
  \centering
  \includegraphics[width=0.99\linewidth]{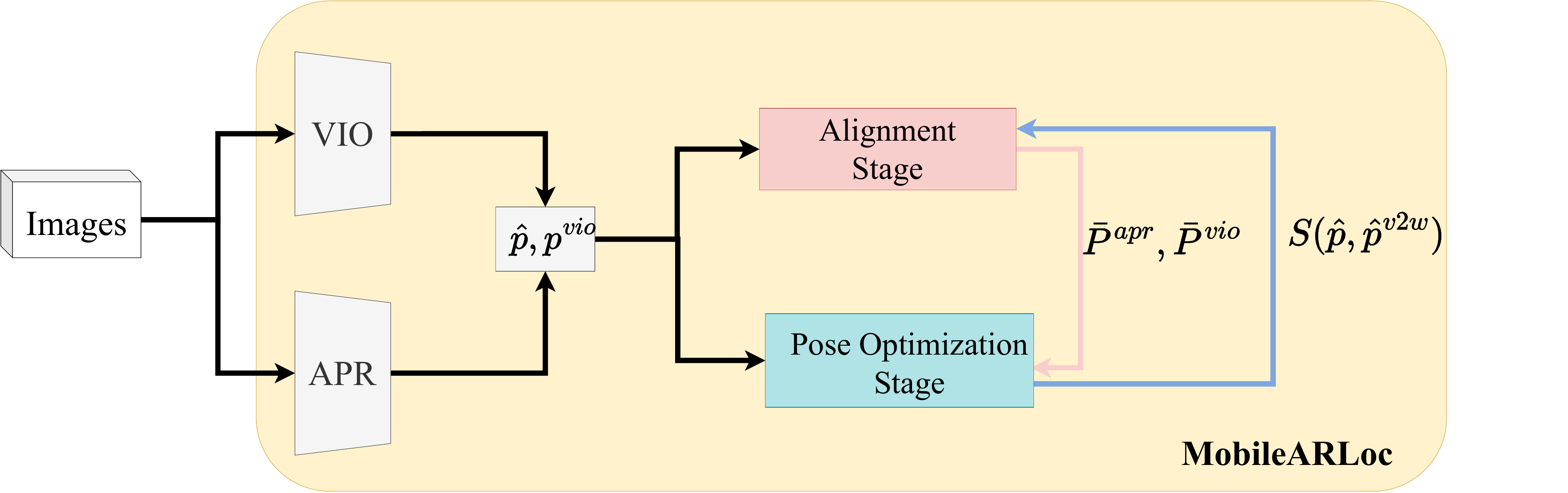}
\caption{ \sysname framework in a mobile AR system.}
\label{fig:framework}
\end{figure}

\subsection{\sysname framework} 
\label{subsec:vio-apr}
Based on the above subsections,  we present \sysname, a new APR-agnostic framework that combines the outputs of APRs and information from smartphones' VIO systems to improve pose prediction accuracy. 
The framework keeps the most reliable prediction poses with the help of VIO, identifies unreliable poses, and optimizes them based on reliable poses. 
\sysname combines the distinctive features of APR predictions, which are locally noisy but drift-free, with mobile VIO systems, which are locally smooth but tend to drift, as shown by~\cite{brahmbhatt2018geometry}. 
This framework consists of two alternating looping stages: \textit{Alignment} and \textit{Pose optimization}, as shown in Figure~\ref{fig:framework}.  The \textit{Alignment} stage identifies multiple reliable poses to calculate the \textit{reference pose}. The \textit{Pose Optimization} stage optimizes unreliable poses based on this \textit{reference pose} and the VIO poses.
\sysname adaptively goes back to the \textit{Alignment} phase to recalculate the \textit{reference pose} and thus negate the effect of VIO drift.


\textbf{Alignment} stage checks the odometry of consecutive $N+1$ images. 
We consider $I_j$ to be the first image to enter the alignment stage. When all $N$ consecutive pairs of images satisfy the requirement that $\text{RPE}^{<apr,vio>} \leq d_{th}$ and  $\text{ROE}^{<apr,vio>} \leq o_{th}$, predictions $\{\hat{p}_i\}^{j+N+1}_{i=j}$ of these $N+1$ consecutive images from $I_j$ to $I_{j+N+1}$ can considered as accurate predictions.  $d_{th}$ and $o_{th}$ are the relative pose checker's distance and orientation threshold to filter the inaccurate estimated poses. If the difference between $\hat{u}_{i,i+1}$ and $u_{i,i+1}^{vio}$ is less than $d_{th}$ and $o_{th}$ simultaneously,  poses $\hat{p}_i$ and $\hat{p}_{i+1}$ become candidate reliable predictions (RPs). If the difference between $\hat{u}_{i,i+1}$ and $u_{i,i+1}^{vio}$ is larger than $d_{th}$ or $o_{th}$, we assume $\hat{p}_i$ is inaccurate and discard all previous candidate RPs. Upon getting $N+1$ candidate RPs,
We perform geometric averaging from these RPs using Weiszfeld's algorithm. 
and ~\cite{gramkow2001averaging} 
to obtain a \textit{reference pose} $\Bar{P}^{apr}$ since we assume the pose error of APR is normally distributed in space.  We  perform the same geometric averaging from corresponding VIO poses to get the \textit{reference pose} $\Bar{P}^{vio}$ in the VIO system. Once $\Bar{P}^{apr}$ and $\Bar{P}^{vio}$ are obtained,
\sysname moves to the \textit{Pose Optimization} stage. 
     

In \textbf{Pose Optimization} stage, the framework checks the current predicted pose against the previous pose as follows. 
    For the  subsequent predicted poses $\{\hat{p}_i\}_{i = j+N+2}$ and corresponding VIO poses $\{p_i^{vio}\}_{i = j+N+2}$, we get the odometry $\hat{u}_{i-1,i}$ of $\hat{p}_{i-1}$ and $\hat{p}_{i}$ in APR coordinates and odometry $\hat{u}_{i-1,i}^{vio}$ of $p_{i-1}^{vio}$ and $p_i^{vio}$ in VIO coordinates respectively.  If $\text{RPE}^{<apr,vio>}_{i-1,i} \leq d_{th}$ and  $\text{ROE}^{<apr,vio>}_{i-1,i} \leq o_{th}$ is satisfied, the APR output for this image is considered reliable as the difference of odometry between APR and VIO is small. The pose is considered reliable and can be output directly.  Otherwise, the pose is optimized by $OptimizePose(p^{vio},\Bar{P}^{apr},\Bar{P}^{vio})$.
We calculate the rigid transformation between VIO coordinates and world coordinates using $\Bar{P}^{apr}$ and $\Bar{P}^{vio}$. Then we transform $p^{vio}$ to world coordinates as $p^{v2w}$ and replace the unreliable pose.

 \textbf{We call the $N+1$ consecutive predictions in \textit{Alignment} stage for calculating the average reference pose and predictions pass the relative pose threshold in \textit{Pose Optimization} stage as reliable predictions (RPs).}
To compensate for VIO drift, we detect the drift and make the system loop back to the \textit{Alignment} stage adaptively:
for each RP in \textit{Pose Optimization} stage, we calculate the similarity between the $\hat{p} = <\hat{\mathbf{x}}, \hat{\mathbf{q}}>$ and $p^{v2w} = <\mathbf{x}^{v2w},\mathbf{q}^{v2w}>$ from $OptimizePose(p^{vio},\Bar{P}^{apr},\Bar{P}^{vio})$:
\begin{equation}
    S(\hat{p},p^{v2w}) = \frac{(\frac{ \hat{\mathbf{x}} \cdot \mathbf{x}^{v2w} }{||\hat{\mathbf{x}}||_2\cdot||\mathbf{x}^{v2w}||_2} + |\frac{ \hat{\mathbf{q}} \cdot \mathbf{q}^{v2w} }{||\hat{\mathbf{q}}||_2\cdot||\mathbf{q}^{v2w}||_2} |)}{2}
\label{eq:simi_drift}
\end{equation}
where $-0.5 <S(\hat{p},p^{v2w})<1$. We use similarity here to balance the difference of translation and rotation because they have different units. If consecutive $N$ reliable predictions have $S(\hat{p},p^{v2w}) \leq \gamma$, it indicates drift happens and the system loops back to the \textit{Alignment} since $p^{v2w}$ should be very close to RPs. The local tracking of VIO system is used between the last pose in \textit{Pose Optimization} stage until new RPs are found in the \textit{Alignment} stage and new reference poses are updated.

\section{Dataset}
\label{sec:ds}   

We collect a dataset of image and VIO data using iPhone 14 Pro Max, the flagship ARKit 6 phone at the time. 
The resolution of all images is $1920\times1440$. All images are fed into an SfM framework using COLMAP~\cite{schonberger2016structure} to get the GT. We compare the GT with the pose labels of VIO.
Our dataset highlights the low drift of current mobile VIO solutions, such as ARKit, supporting our assumption that VIO system can reinforce absolute pose estimation.   
Table~\ref{tab:ds} provides  summary statistics, as well as the storage requirements and runtime per request for HLoc, PN, and MS-T. (see Section~\ref{subsec:eff}). 

 \begin{table}[t]
\caption{Dataset details and statistics for HLoc (Image Retrieval with top-20 recall), PN, and MS-T.} 
\centering
\setlength{\tabcolsep}{1pt}
\resizebox{\columnwidth}{!}{
\begin{tabular}{c|c|cc|c|ccc|ccc} 
& Scenes & \multicolumn{2}{c|}{ Dataset quantity } & \text { Spatial } & \multicolumn{3}{c|}{Storage (MB)} & \multicolumn{3}{c}{Runtime (\,ms)} \\
& & Train  & Test  &  Extent (m) & HLoc & PN & MS-T & HLoc  &PN & MS-T\\
\hline 
&Square  & 2058 & 1023&  40$\times$25 &6500&85&71&4965&4&15\\
Outdoor&\text { Church } & 1643 & 853  &  50$\times$40&6600&85&71&6659&4&15\\
& Bar  & 1834 & 838 & 55$\times$ 35&6600&85&71&6230&4&15\\
\hline 
&Stairs  &  873& 222 &  5.5 $\times$ 4.5 $\times$ 6 &1500&85&71&2263&4&15\\
Indoor&\text {Office } &  1479 &635  & 7.5 $\times$ 4 &2300&85&71&5722&4&15\\
& Atrium  & 1694 & 441 &  30 $\times$ 50&5100&85&71&4500&4&15\\
\hline 
\end{tabular}
}
\label{tab:ds}
\end{table}

\section{Implementation}

\subsection{Desktop Implementation}
\label{subsec:desktop}

There are four hyperparameters. $d_{th}$ and $o_{th}$ are the relative pose checker's distance and orientation thresholds to filter inaccurate pose estimations.  
If among $N+1$ consecutive images, $N$ image pairs pass the relative pose checker, we consider these pose estimations to be accurate.  We set $\gamma = 0.99$, $d_{th} = 0.4m$ and $o_{th} = 4^ \circ$ over all datasets. We set $N = 2$, with one frame processed per second for the AR application and all the experiments in this paper. We implement our framework over two APR models:

\noindent\textbf{PN}. PoseNet (PN) is the baseline method. Since there is no open source code for PoseNet~\cite{kendall2015posenet}, we  follow~\cite{kendall2017geometric, brahmbhatt2018geometry} and use ResNet34~\cite{he2016deep} as the backbone network. 


\noindent\textbf{MS-T}. MS-Transformer~\cite{shavit2021learning} (MS-T) extends the single-scene paradigm of APR to learning multiple scenes in parallel and is one of the most recent APRs with official opensource code. Therefore, we trained one MS-T for all outdoor scenes and one MS-T for all indoor scenes using the official code\footnote{\url{https://github.com/yolish/multi-scene-pose-transformer}}.

\textbf{We note APR methods integrated into our \sysname framework as $\text{APR}^{vio}$}. During training, all input images are resized to $256\times 256$ and then randomly cropped to $224\times 224$. For both PN and MS-T, we set an initial learning rate of $\lambda= 10^{-4}$.  
All experiments for evaluation in Section~\ref{sec:exp} are performed on an NVIDIA GeForce GTX 3090 GPU.

\subsection{Application Implementation}
\label{subsec:mobileapp}
We implement \sysname as a mobile AR app using Unity and ARKit to run on an iPhone 14 Pro Max.  We convert the pre-trained PN to ONNX format and incorporated it into a Unity application. We use OpenCVforUnity\footnote{\url{https://assetstore.unity.com/packages/tools/integration/opencv-for-unity-21088}} 
for processing query images and use Barracuda transferring resized images to tensor as the input of the network.

\section{System Performance}


\subsection{Desktop Implementation}
\label{subsec:eff}

As mentioned in Section~\ref{sec:intro}, structure-based methods tend to require significant resources that are not available on mobile devices. Table~\ref{tab:ds} shows the performance of HLoc~\cite{sarlin2019coarse} pipelines\footnote{\url{https://github.com/cvg/Hierarchical-Localization}} prevents one-device camera relocalisation. In \sysname, the APR only requires storing neural network weights. The memory requirement of the APR thus remains constant, between 71 and 85\,MB depending on the backbone model. Meanwhile, HLoc pipelines require 1) a pre-built 3D model; 
2) an image database; 3) a local descriptor database; and 4) the models for image retrieval and feature extraction. The memory represents between 1.5 and 6.6\,GB per scene. 
APRs only require a single forward pass for each query image, leading to a runtime between 4 and 15\,ms. In contrast, HLoc pipelines take up to 6.7\,s on larger scenes.  

\subsection{Mobile Implementation}

We assess the performance of our framework on an iPhone 14 Pro Max device on the setup described in Section~\ref{subsec:mobileapp}. We measure each parameter over 200 samples. The average processing time per image is 37\,ms while the average time for PN to infer an image is 39.5\,ms. Under current ARKit's implementation, VIO runs in a parallel loop every frame. 
Therefore, our pipeline can perform absolute camera localization on a mobile device in less than 80\,ms. 
The ResNet34-based PN requires only 85 MB for weight storage.  


%% file: Sections/4Experiment.tex
\section{Dataset Evaluation}

\label{sec:exp}
We evaluate our framework on the datasets described in Section~\ref{sec:ds} and Section~\ref{subsec:desktop}.  Due to the disparity in computational scale between our approach and structure-based methods like HLoc (refer to Table~\ref{tab:ds}), we do not present results for HLoc. 

\begin{table}[t]
\caption{Mean and median absolute translation/rotation errors in $m/^\circ$. The ratio of RPs and Opt. poses are provided in Table~\ref{tab:ratio}. }
\centering
\setlength{\tabcolsep}{1pt}
\setlength{\tabcolsep}{1pt}
\resizebox{\columnwidth}{!}{
\begin{tabular}{l|cc|cc|cc|cc}
\hline
\multicolumn{5}{c}{Only RPs Mean} &\multicolumn{4}{c}{Only RPs Median}\\
\hline
 \textbf{Outdoor} &  PN &  PN$^{vio}$(ours) &  MS-T &  MS-T$^{vio}$(ours) &  PN &  PN$^{vio}$(ours) &  MS-T &  MS-T$^{vio}$(ours) \\ \hline 
     Square & 2.1/7.07&\textbf{1.17/3.57}& 2.50/4.14 &\textbf{1.64/2.46} & 1.11/3.61 & \textbf{0.76/3.04}&1.5/2.14&\textbf{0.81/1.78}\\
Church& 1.53/8.04 &\textbf{0.82/4.0}&1.91/11.9& \textbf{0.65/3.0} & 0.73/3.97 & \textbf{0.59/3.0} &0.71/3.08& \textbf{0.49/2.01} \\
Bar & 1.44/4.03&\textbf{0.78/2.89}&1.65/2.82&\textbf{0.86/1.97}& 0.66/2.82 & \textbf{0.52/2.38}&0.69/1.65&\textbf{0.52/1.47}\\
 \hline
 average & 1.69/6.38  & \textbf{0.92/3.49} & 2.02/6.29 & \textbf{1.05/2.48}  & 0.83/3.47& \textbf{0.62/2.81} &0.97/2.29 & \textbf{0.61/1.75}\\\hline
\multicolumn{5}{c}{Only Opt. Mean} &\multicolumn{4}{c}{Only Opt. Median}\\
\hline
Square  &2.8/10.5&\textbf{1.4/2.14}&3.25/5.54&\textbf{1.27/2.12} &1.45/4.38&\textbf{0.98/2.16}&2.22/2.44&\textbf{1.04/1.89} \\
Church &2.07/11.2 &\textbf{1.08/1.98} &2.85/18.6&\textbf{1.1/1.86} &0.93/5.04 & \textbf{0.89/1.87}&1.13/4.74&\textbf{0.93/1.70} \\
Bar &2.18/5.35&\textbf{1.07/2.12} &2.42/3.74&\textbf{0.96/1.56}&0.94/3.55 &\textbf{0.86/2.18} &0.99/1.75&\textbf{0.86/1.35} \\
 \hline
average & 2.35/9.02 & \textbf{1.18/2.08}  &  2.84/9.29 & \textbf{1.1/1.85} & 1.11/4.32 & \textbf{0.91/2.07} & 1.45/2.98 & \textbf{0.94/1.65}\\\hline
\multicolumn{5}{c}{RPs + Opt. Mean} &\multicolumn{4}{c}{RPs+ Opt. Median}\\
\hline
Square  & 2.1/7.07  &  \textbf{1.3/2.8} & 2.49/4.14& \textbf{1.44/2.27} & 1.11/3.61 & \textbf{0.84/2.36} & 1.5/2.14& \textbf{0.96/1.87}\\
Church &  1.53/8.04 & \textbf{0.96/2.88}& 1.9/11.9 & \textbf{0.9/2.34}&   \textbf{0.73}/3.97 & 0.74/\textbf{2.13}&\textbf{0.71}/3.08& 0.73/\textbf{1.79}\\
Bar & 1.44/4.03 & \textbf{0.92/2.53}  &1.65/2.82&\textbf{0.91/1.78}&\textbf{0.66}/2.82  & 0.69/\textbf{2.26} &\textbf{0.69}/1.65& 0.7/\textbf{1.44} \\
 \hline
average &1.69/6.38  & \textbf{1.06/2.74}  & 2.02/6.29 &  \textbf{1.08/2.13} &0.83/3.47 & \textbf{0.76/2.25} & 0.97/2.29 & \textbf{0.8/1.7} \\\hline
\hline
\multicolumn{5}{c}{Only RPs Mean} &\multicolumn{4}{c}{Only RPs Median}\\
\hline
 \textbf{Indoor} &  PN &  PN$^{vio}$(ours) &  MS-T &  MS-T$^{vio}$(ours) &  PN &  PN$^{vio}$(ours) &  MS-T &  MS-T$^{vio}$(ours) \\ \hline 
Stairs & 0.35/7.51& \textbf{0.27/5.25}&0.26/6.8&\textbf{0.22/4.56} & 0.27/4.89& \textbf{0.25/4.88} &0.18/4.33&\textbf{0.16/3.6}\\
Office& 0.55/11.7 &\textbf{0.45/7.3} &0.48/13.3& \textbf{0.40/7.36} & 0.36/6.21 & \textbf{0.32/5.55} &0.35/4.66& \textbf{0.30/4.11}\\
Atrium & 2.71/10.2& \textbf{1.79/6.14}&3.66/9.08&\textbf{1.95/5.03} & 1.78/5.54& \textbf{1.2/4.3} &2.0/3.98&\textbf{1.35/3.23}\\
 \hline
 average & 1.2/9.8 &  \textbf{0.84/6.23} & 1.47/9.73 & \textbf{0.86/5.65} & 0.80/5.55 & \textbf{0.59/4.91} &0.84/4.32 & \textbf{0.6/3.65}\\\hline
\multicolumn{5}{c}{Only Opt. Mean} &\multicolumn{4}{c}{Only Opt. Median}\\
\hline
Stairs  &0.50/11.76 & \textbf{0.21/4.88}  &0.32/11.9&\textbf{0.19/3.5} & 0.40/4.98& \textbf{0.20/4.96}&0.25/7&\textbf{0.13/3.74}\\
Office & 0.9/25.2 &\textbf{0.51/6.49} &0.8/31.3 & \textbf{0.33/4.27} & 0.79/12.2&\textbf{0.36/6.1}&0.6/12.8 &\textbf{0.21/3.34}\\
Atrium  & 3.13/13.2& \textbf{2.26/12.9} &4.6/13.5&\textbf{1.2/1.95} & 2.53/7.3& \textbf{2.09/4.9} &2.6/6& \textbf{1.16/1.6} \\
 \hline
average & 1.51/16.73  & \textbf{0.99/8.09}  & 1.91/18.9 &  \textbf{0.57/3.24} & 1.24/8.16 & \textbf{0.88/5.32} & 1.15/8.6 &  \textbf{0.5/2.89}\\\hline
\multicolumn{5}{c}{RPs + Opt. Mean} &\multicolumn{4}{c}{RPs+ Opt. Median}\\
\hline
Stairs & 0.35/7.51& \textbf{0.27/5.25} &0.26/6.8&\textbf{0.22/4.56} &0.27/\textbf{4.89} &\textbf{0.22}/4.90&0.18/4.33&\textbf{0.15/3.68}\\
Office & 0.55/11.7 & \textbf{0.47/7.13}& 0.48/13.3 & \textbf{0.39/6.89} &0.36/6.21  &\textbf{0.34/5.8}& 0.35/4.66& \textbf{0.28/4.0}\\
Atrium & 2.71/10.2 & \textbf{1.99/9.0}  & 3.66/9.08 & \textbf{1.59/3.52} & 1.78/5.54  &\textbf{1.57/4.81} &2.0/3.98& \textbf{1.21/2.61}\\
 \hline
average &1.2/9.8  &  \textbf{0.91/7.13} & 1.47/9.73 &  \textbf{0.73/4.99} &0.80/5.55 & \textbf{0.71/5.17} &0.84/4.32 & \textbf{0.55/3.43} \\\hline
\end{tabular}
}
\label{tab:ape_kopt}
\end{table}

  We evaluate the performance of APR and our framework through two primary metrics. We consider the mean and median APE and AOE in Tables~\ref{tab:ape_kopt} as shown in Equation (\ref{eq:ape_apr}) and Equation (\ref{eq:aoe_apr}) for all test frames. We also evaluate the percentage of test images with pose predicted with high ($0.25m, 2^\circ$), medium ($0.5m, 5^\circ$), and low ($5m, 10^\circ$) accuracy levels proposed by~\cite{sattler2018benchmarking} in Table~\ref{tab:ratio}. The higher the percentage of each accuracy level, the better the performance.

 \begin{equation}
    \text{APE}^{<apr, GT>}  = || \mathbf{\hat{x}}_i - \mathbf{x}_i||_2
\label{eq:ape_apr}
\end{equation}

\begin{equation}
    \text{AOE}^{<apr, GT>}  = 2\arccos{|\mathbf{q}_{i}^{-1}\mathbf{\hat{q}}_{i}|}\frac{180}{\pi}
\label{eq:aoe_apr}
\end{equation}

\begin{figure}[t!]
 \centering
   \subfigure[PN and PN$^{vio}$]{
   \centering
 \label{fig:subfig:f} 
 \includegraphics[width=.48\linewidth]{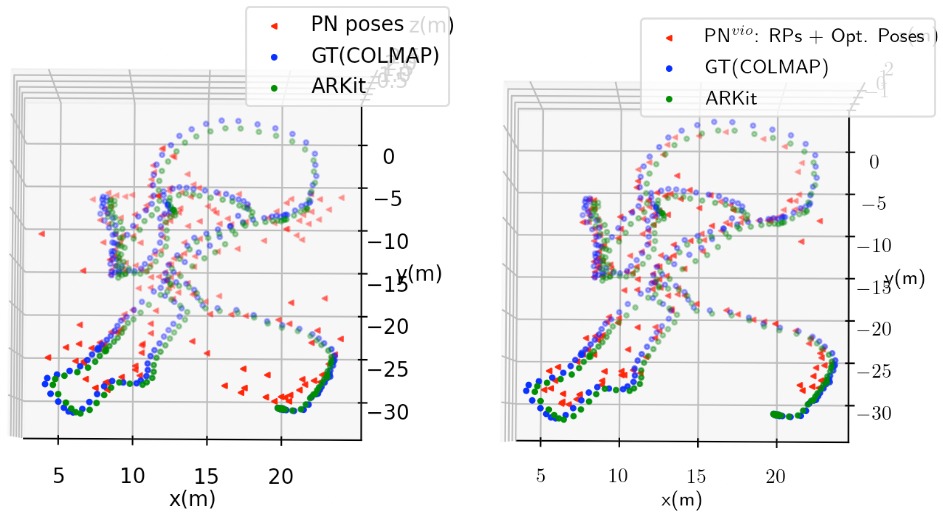}}
 \subfigure[MS-T and MS-T$^{vio}$]{
 \centering
 \label{fig:subfig:a} 
 \includegraphics[width=.48\linewidth]{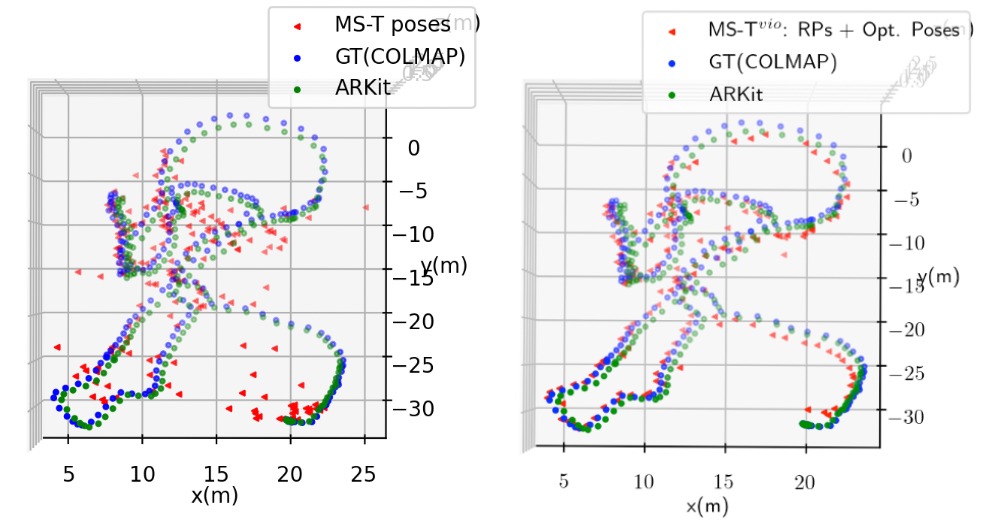}}
 \caption{Pose predictions for APR (left) and $\text{APR}^{vio}$ (right) for one test sequence in the Square scene.
 Using \sysname significantly decreases the number of predictions with large error as well as the noisiness of pose estimation error compared to APR alone.
 }
 \label{fig:track} 
\end{figure}

\begin{figure}[t!]
 \centering
   \subfigure[PN and PN$^{vio}$ (Trans.)]{
   \centering
 \label{fig:subfig:f} 
 \includegraphics[width=.45\linewidth]{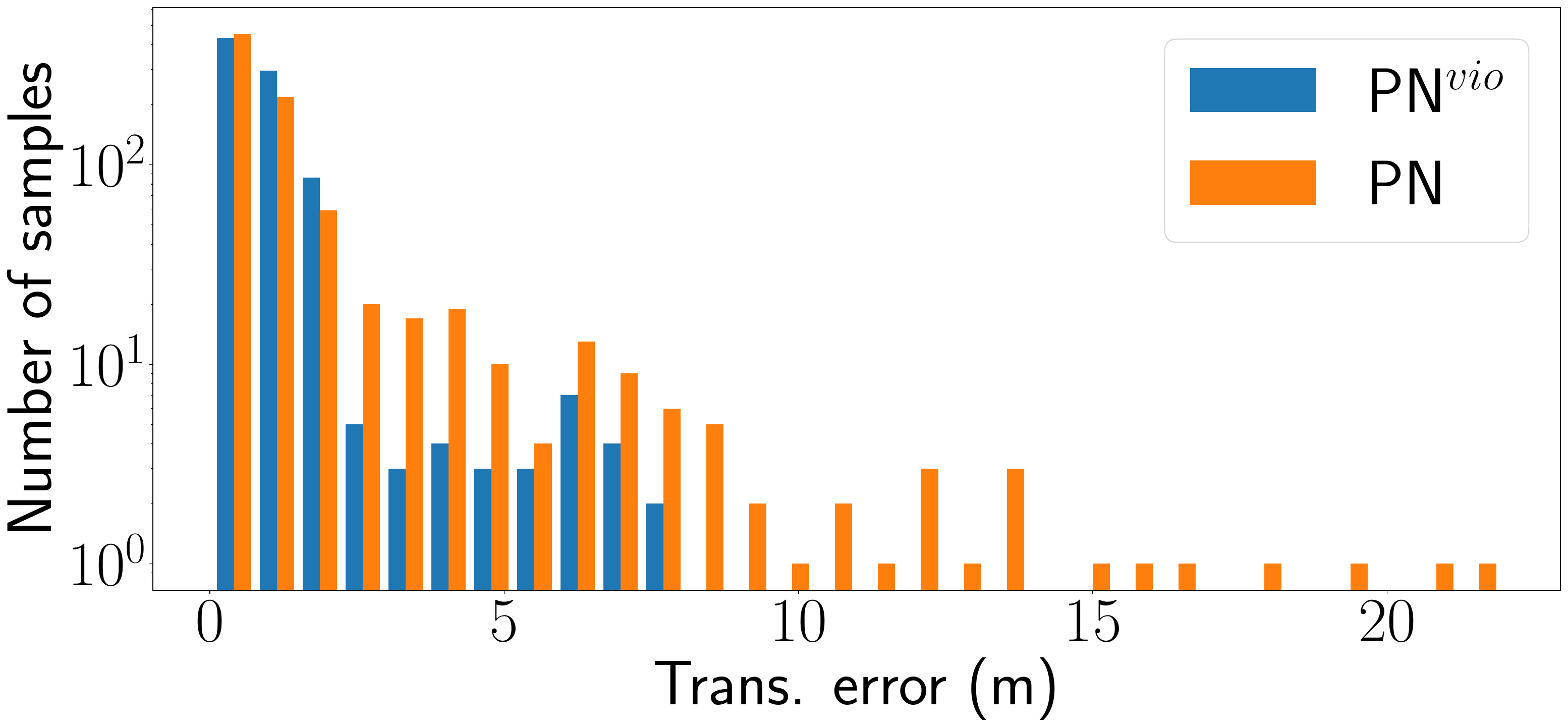}}
 \subfigure[MS-T and MS-T$^{vio}$  (Trans.)]{
 \centering
 \label{fig:subfig:a} 
 \includegraphics[width=.45\linewidth]{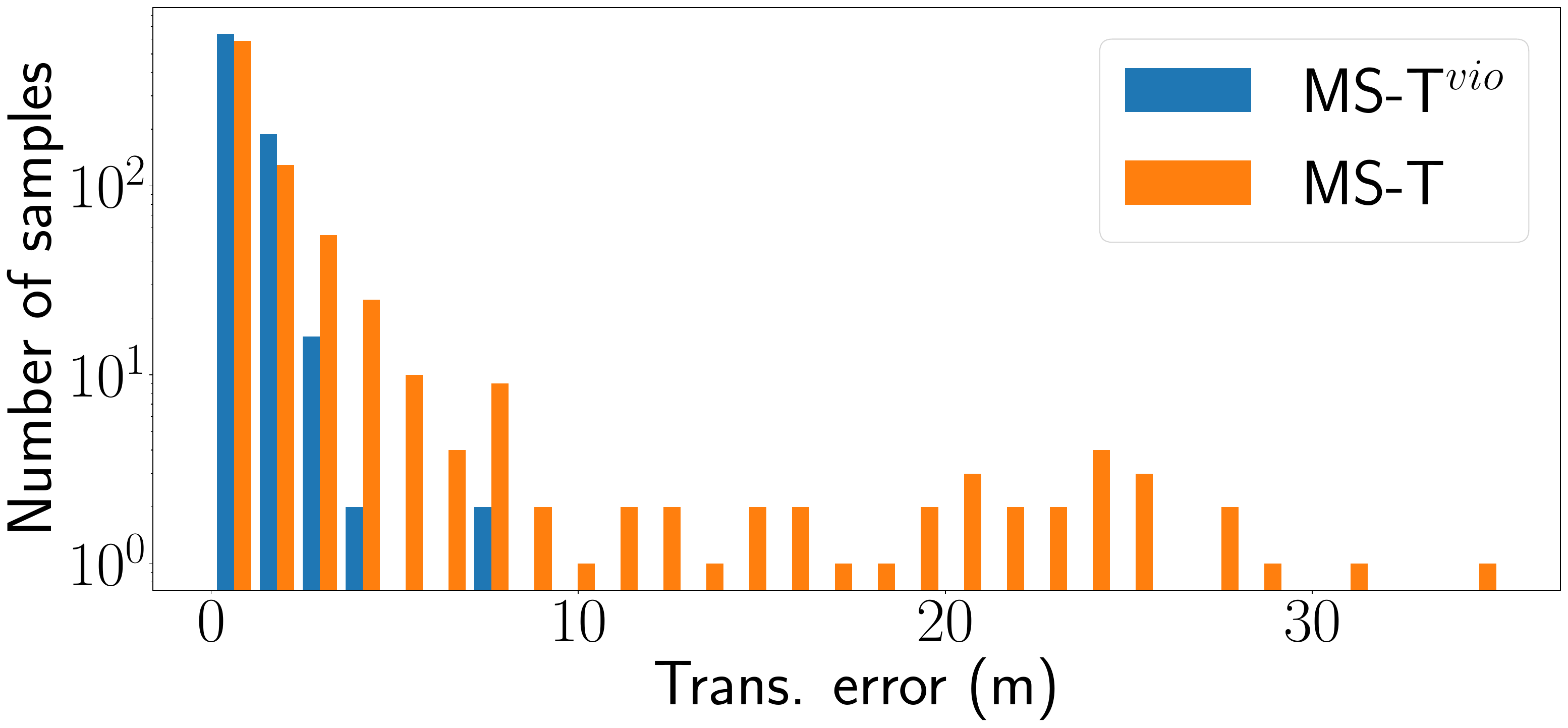}}
 \caption{Pose error distribution for Church scene.
Using \sysname significantly decreases the number of predictions with large error.
 }
 \label{fig:dist} 
\end{figure}

\subsection{Results}

\noindent\textbf{Reliable predictions:} 
For outdoor scenes, Table~\ref{tab:ratio} demonstrates that selecting RPs comprising 41.8\% to 51.3\% of total PN predictions and 41.6\% to 51\% of total MS-T predictions significantly increases the percentage of each accuracy level. Consequently, pose estimates below the low accuracy threshold (5m, 10$^\circ$) are greatly reduced in all three outdoor scenes. Moreover, RPs selected by PN$^{vio}$ and MS-T$^{vio}$ exhibit higher mean and median accuracy across all predictions compared to PN and MS-T (Table~\ref{tab:ape_kopt}). The mean accuracy improvement surpasses the median accuracy improvement. In the three outdoor scenes, PN$^{vio}$-preserved RPs reduce mean translation error by 44\% to 46\% and median translation error up to 33\%. Mean rotation accuracy improves by 28\% to 50\%, while median rotation accuracy improves up to 22\%. MS-T$^{vio}$-preserved RPs reduce mean translation error by 34\% to 66\% and median translation error by 11\% to 32\%. Mean rotation accuracy improves by 30\% to 75\%, and median rotation error improves by 11\% to 35\%. The improvement for indoor scenes follows a similar pattern. Table~\ref{tab:ratio} shows that the percentage of each accuracy level is greatly increased by selecting the RPs that amount for 48.5\% to 76.2\% of the total PN pose estimations and 46.3\% to 78.7\% of the total MS-T pose estimations. APE$^{<apr,GT>}$ and AOE$^{<apr,GT>}$ are greatly reduced as shown in Table~\ref{tab:ape_kopt}. These outcomes indicate that a significant portion of predictions with large errors contribute to the lower accuracy. Our framework effectively identifies RPs using VIO.

\noindent\textbf{Optimized Poses:} 
Tables~\ref{tab:ape_kopt} and~\ref{tab:ratio} provide confirmation that a portion of the pose estimates that fail the relative pose checker in the \textit{pose optimization} stage exhibit larger errors compared to the median and mean pose errors of all predictions. The difference in accuracy is even more pronounced when compared to the selected RPs. These unreliable poses significantly impact overall accuracy. For outdoor scenes, PN$^{vio}$ improves mean translation accuracy by 48\% to 51\% and median translation accuracy by up to 32\%. It also enhances mean rotation accuracy by 60\% to 82\% and median rotation accuracy by 39\% to 63\%. Similarly, MS-T$^{vio}$ improves mean translation accuracy by 60\% to 61\% and median translation accuracy by 13\% to 53\%. It also enhances mean rotation accuracy by 58\% to 90\% and median rotation accuracy by 23\% to 64\%. 
Pose estimates below the low accuracy level are significantly reduced for both PN$^{vio}$ and MS-T$^{vio}$. In the Church and Bar scenes, no optimized pose estimates of MS-T$^{vio}$ exceed 5 meters and 10 degrees. For indoor scenes, 20.2\% and 36.7\% of PN pose estimates and 14.2\% to 44.7\% of the MS-T pose estimates that do not pass the relative pose threshold in the \textit{pose optimization} stage have larger error compared with the median and mean pose error of all predictions. The optimized accuracy has experienced a substantial improvement.

\noindent\textbf{Total results:}
As shown in Table~\ref{tab:ratio}, less than 10 percent of the predictions in test set are filtered out as unreliable predictions in \textit{alignment} stage except for atrium. \sysname effectively optimizes unreliable predictions in \textit{pose optimization} stage, which leads to a much higher mean accuracy than original APRs on both translation and rotation for all three scenes. 
Figure~\ref{fig:track} and Figure~\ref{fig:dist} show \sysname improves accuracy by reducing the incidence of outliers with large error and the noisiness of APR predictions.

\begin{table}[hbt!]
\caption{Percentage (\%) of RPs, optimized poses and total poses predicted with high (0.25m, $2^{\circ}$), medium (0.5m, $5^{\circ}$), and low (5m, $10^{\circ}$) accuracy~\cite{sattler2018benchmarking}  (higher is better). The value in parentheses represents the ratio (\%) of RPs, opt. poses, and RPs + opt. poses in the test set.}
\setlength{\tabcolsep}{1pt}
\resizebox{\columnwidth}{!}{
\begin{tabular}{c|l|cr|cr}
\hline
\multicolumn{6}{c}{Only Reliable Predictions}\\
\hline
Dataset &Scenes  & PN & PN$^{vio}$(ours)  & MS-T & MS-T$^{vio}$(ours)  \\ \hline
 &Square  & 3.2/18.4/87 & \textbf{5.6/27.2/96.4} (41.8) &2.9/17.9/81.1& \textbf{3.8/26.9/91.9} (43.2)\\
Outdoor & Church & 2.2/21.9/82.1& \textbf{3.8/33.4/93.0} (43.5) &6.9/29.7/79.0& \textbf{12.4/48.7/96.1} (41.6)\\
&Bar  &  4.1/31.7/89.5 & \textbf{6/43/96.7} (51.3) &8.5/34.6/90.8 &\textbf{12.2/46.4/97.4} (51)\\
 \hline
\multicolumn{6}{c}{Only Optimization}\\
\hline
&Square  & 1.8/13.7/79 (49.9) & \textbf{4.1/20.8/94.3} (49.9)& \textbf{2}/10.1/72 (53.1) & 1.5/\textbf{12.7/97.8}(53.1)\\
Outdoor&Church& 1.1/12.7/73.6 (55.6) & \textbf{3.6/21.9/97.5} (55.6)  & \textbf{3.1}/15.3/66.3 (57.4) & 1.8/\textbf{15.7/100} (57.4) \\
&Bar  &  1.8/\textbf{19.5}/81 (45.9) & \textbf{3.1}/16.8/\textbf{100} (45.9) & \textbf{4.8/23.0}/85.3 (44.6)& 4/15.8/\textbf{100} (44.6)\\
 \hline
 \multicolumn{6}{c}{Reliable Predictions + Optimization}\\
 \hline
 &Square  & 3.2/18.4/87.0 & \textbf{4.8/23.7/95.3} (91.7) &\textbf{2.9}/17.9/81.1& 2.5/\textbf{19.1/95.1} (96.3)\\
Outdoor & Church & 2.2/21.9/82.1 & \textbf{3.7/27.0/95.5} (99.1)&\textbf{6.9/29.7}/79.0& 6.3/29.6/\textbf{98.3} (97)\\
&Bar  &  4.1/\textbf{31.7}/89.5 & \textbf{4.7}/30.7/\textbf{98.3} (97.2)  &\textbf{8.5/34.6}/90.8& 8.4/32.1/\textbf{98.6} (92)\\
 \hline
 \hline
\multicolumn{6}{c}{Only Reliable Predictions}\\
\hline
Dataset &Scenes  & PN & PN$^{vio}$(ours)  & MS-T & MS-T$^{vio}$(ours)  \\ \hline
 &Stairs   & 5.4/48.2/87.4& \textbf{7.7/51.4/93.7} (64) &18.5/58.1/86.9& \textbf{22.9/69.3/94.1} (68.9)\\
Indoor&Office & 3.8/31.0/72.8 &\textbf{4.3/36/80.8} (76.2) &7.1/46.0/80& \textbf{8.8/54.7/90.2} (78.7)\\
&Atrium  & \textbf{0}/5.4/71 & \textbf{0}/\textbf{8.4/86} (48.5) &0.4/7.3/66.7&\textbf{1/13.7/88.7} (46.3)\\
 \hline
 \multicolumn{6}{c}{Only Optimization}\\
 \hline
 &Stairs   & \textbf{1.3}/42.9/75.3 (36.7)  & 0\textbf{/53.2/100} (36.7) &9.1/34.8/71.2 (29.7) & \textbf{34.8/69.7/100} (29.7)\\
Indoor&Office &1.6/14.8/46.9 (20.2) &\textbf{9.4/18/89.8} (20.2) & 1.1/15.6/41.1 (14.2) & \textbf{23.3/51.1/94.4} (14.2)\\
&Atrium  & \textbf{0}/2.5/64.6 (35.8)  & \textbf{0}/\textbf{5.1/75.3} (35.8)  &0/1.5/46.7 (44.7) & \textbf{0.5/4.1/100} (44.7)\\
\hline
\multicolumn{6}{c}{Reliable Predictions + Optimization}\\
\hline
 &Stairs   & \textbf{5.4}/48.2/87.4 & 5/\textbf{52.1/95.9} (90.7) &18.5/58.1/86.9& \textbf{26.5/69.4/95.9} (98.6)\\
Indoor&Office & 3.8/31.0/72.8  & \textbf{5.4/32.2/82.7} (96.4) &7.1/46.0/80& \textbf{11/54.1/90.8} (92.9)\\
&Atrium  & \textbf{0}/5.4/71.0 & \textbf{0}/\textbf{6.9/81.5} (84.3)  &0/7.3/66.7 & \textbf{0.7/9/94.3} (91)\\
 \hline
\end{tabular}
}
\label{tab:ratio}
\end{table}

\subsection{Analysis}
Table~\ref{tab:ratio} show that PN and MS-T have predictions that are very inaccurate with large errors more than 5 meters and 10 degrees in both outdoor and indoor scenes. By calculating the reference pose with RPs identified by VIO system, unreliable predictions are optimized, resulting in significant improvement.
Outdoor,  our framework improves the accuracy of $\text{MS-T}$ by $47\%$ on mean translation error and $66\%$ on mean rotation error over all scenes average.  
Indoor, it improves the accuracy of $\text{MS-T}$ by $50\%$ on mean translation error and $49\%$ on mean rotation error on all scenes average. 
Compared to the median accuracy, our method provides a greater improvement in mean accuracy. This is because the accuracy of our optimization depends on the accuracy of the reference pose, which is basically about the same error as the median accuracy. Outdoor datasets yield better results due to the lower accuracy of the APR in indoor scenes. This leads to misclassification of inaccurate poses during \textit{alignment} (case (3) in Section~\ref{sec:detect}), making the calculation of the reference pose challenging and introducing further inaccuracies during \textit{pose optimization}.

%% file: Sections/6Conclusion.tex
\section{Conclusion}
This paper introduces \sysname, a framework that combines an APR with a local VIO tracking system to improve the accuracy and stability of localization for markerless mobile AR. 
The VIO evaluates and optimizes the APR's accuracy while the APR corrects VIO drift, resulting in improved positioning. 
We evaluate \sysname through dataset simulations.  \sysname improves the position accuracy by up to 50\% and rotation by up to 66\% for different APRs. The mobile app can perform pose estimation in less than 80\,ms with minimal storage and energy consumption. Due to its low system footprint, high accuracy, and robustness, \sysname enables pervasive markerless mobile AR at a large scale.


%% file: main.bbl
\begin{thebibliography}{10}
\providecommand{\url}[1]{#1}
\csname url@samestyle\endcsname
\providecommand{\newblock}{\relax}
\providecommand{\bibinfo}[2]{#2}
\providecommand{\BIBentrySTDinterwordspacing}{\spaceskip=0pt\relax}
\providecommand{\BIBentryALTinterwordstretchfactor}{4}
\providecommand{\BIBentryALTinterwordspacing}{\spaceskip=\fontdimen2\font plus
\BIBentryALTinterwordstretchfactor\fontdimen3\font minus \fontdimen4\font\relax}
\providecommand{\BIBforeignlanguage}[2]{{%
\expandafter\ifx\csname l@#1\endcsname\relax
\typeout{** WARNING: IEEEtran.bst: No hyphenation pattern has been}%
\typeout{** loaded for the language `#1'. Using the pattern for}%
\typeout{** the default language instead.}%
\else
\language=\csname l@#1\endcsname
\fi
#2}}
\providecommand{\BIBdecl}{\relax}
\BIBdecl

\bibitem{dusmanu2019d2}
M.~Dusmanu, I.~Rocco, T.~Pajdla, M.~Pollefeys, J.~Sivic, A.~Torii, and T.~Sattler, ``D2-net: A trainable cnn for joint description and detection of local features,'' in \emph{ieee/cvf conference on computer vision and pattern recognition}, 2019, pp. 8092--8101.

\bibitem{sarlin2019coarse}
P.-E. Sarlin, C.~Cadena, R.~Siegwart, and M.~Dymczyk, ``From coarse to fine: Robust hierarchical localization at large scale,'' in \emph{IEEE/CVF Conference on Computer Vision and Pattern Recognition}, 2019.

\bibitem{taira2018inloc}
H.~Taira, M.~Okutomi, T.~Sattler, M.~Cimpoi, M.~Pollefeys, J.~Sivic, T.~Pajdla, and A.~Torii, ``Inloc: Indoor visual localization with dense matching and view synthesis,'' in \emph{IEEE Conference on Computer Vision and Pattern Recognition}, 2018, pp. 7199--7209.

\bibitem{noh2017large}
H.~Noh, A.~Araujo, J.~Sim, T.~Weyand, and B.~Han, ``Large-scale image retrieval with attentive deep local features,'' in \emph{IEEE international conference on computer vision}, 2017, pp. 3456--3465.

\bibitem{moreau2023imposing}
A.~Moreau, T.~Gilles, N.~Piasco, D.~Tsishkou, B.~Stanciulescu, and A.~de~La~Fortelle, ``Imposing: Implicit pose encoding for efficient visual localization,'' in \emph{IEEE/CVF Winter Conference on Applications of Computer Vision}, 2023, pp. 2892--2902.

\bibitem{braud2020multipath}
T.~Braud, Z.~Pengyuan, J.~Kangasharju, and H.~Pan, ``Multipath computation offloading for mobile augmented reality,'' in \emph{2020 IEEE International Conference on Pervasive Computing and Communications (PerCom)}.\hskip 1em plus 0.5em minus 0.4em\relax IEEE, 2020, pp. 1--10.

\bibitem{fernandez2022implementing}
C.~B. Fernandez, T.~Braud, and P.~Hui, ``Implementing gdpr for mobile and ubiquitous computing,'' in \emph{23rd Annual International Workshop on Mobile Computing Systems and Applications}, 2022.

\bibitem{shavit2021learning}
Y.~Shavit, R.~Ferens, and Y.~Keller, ``Learning multi-scene absolute pose regression with transformers,'' in \emph{IEEE/CVF International Conference on Computer Vision}, 2021, pp. 2733--2742.

\bibitem{sattler2019understanding}
T.~Sattler, Q.~Zhou, M.~Pollefeys, and L.~Leal-Taixe, ``Understanding the limitations of cnn-based absolute camera pose regression,'' in \emph{IEEE/CVF conference on computer vision and pattern recognition}, 2019.

\bibitem{scargill2022here}
T.~Scargill, G.~Premsankar, J.~Chen, and M.~Gorlatova, ``Here to stay: A quantitative comparison of virtual object stability in markerless mobile ar,'' in \emph{2022 2nd International Workshop on Cyber-Physical-Human System Design and Implementation (CPHS)}.\hskip 1em plus 0.5em minus 0.4em\relax IEEE, 2022, pp. 24--29.

\bibitem{brahmbhatt2018geometry}
S.~Brahmbhatt, J.~Gu, K.~Kim, J.~Hays, and J.~Kautz, ``Geometry-aware learning of maps for camera localization,'' in \emph{IEEE conference on computer vision and pattern recognition}, 2018.

\bibitem{kendall2015posenet}
A.~Kendall, M.~Grimes, and R.~Cipolla, ``Posenet: A convolutional network for real-time 6-dof camera relocalization,'' in \emph{IEEE international conference on computer vision}, 2015.

\bibitem{kendall2017geometric}
A.~Kendall and R.~Cipolla, ``Geometric loss functions for camera pose regression with deep learning,'' in \emph{IEEE conference on computer vision and pattern recognition}, 2017, pp. 5974--5983.

\bibitem{chen2022dfnet}
S.~Chen, X.~Li, Z.~Wang, and V.~A. Prisacariu, ``Dfnet: Enhance absolute pose regression with direct feature matching,'' in \emph{ECCV 2022. Tel Aviv, Israel, October 23--27, 2022, Part X}.\hskip 1em plus 0.5em minus 0.4em\relax Springer, 2022.

\bibitem{kendall2016modelling}
A.~Kendall and R.~Cipolla, ``Modelling uncertainty in deep learning for camera relocalization,'' in \emph{2016 IEEE international conference on Robotics and Automation (ICRA)}.\hskip 1em plus 0.5em minus 0.4em\relax IEEE, 2016.

\bibitem{huang2019prior}
Z.~Huang, Y.~Xu, J.~Shi, X.~Zhou, H.~Bao, and G.~Zhang, ``Prior guided dropout for robust visual localization in dynamic environments,'' in \emph{IEEE/CVF international conference on computer vision}, 2019.

\bibitem{moreau2022coordinet}
A.~Moreau, N.~Piasco, D.~Tsishkou, B.~Stanciulescu, and A.~de~La~Fortelle, ``Coordinet: uncertainty-aware pose regressor for reliable vehicle localization,'' in \emph{IEEE/CVF Winter Conference on Applications of Computer Vision}, 2022.

\bibitem{bui20206d}
M.~Bui, T.~Birdal, H.~Deng, S.~Albarqouni, L.~Guibas, S.~Ilic, and N.~Navab, ``6d camera relocalization in ambiguous scenes via continuous multimodal inference,'' in \emph{ECCV 2020: 16th European Conference, Glasgow, UK, August 23--28, 2020, Part XVIII 16}.\hskip 1em plus 0.5em minus 0.4em\relax Springer, 2020.

\bibitem{10160466}
F.~Zangeneh, L.~Bruns, A.~Dekel, A.~Pieropan, and P.~Jensfelt, ``A probabilistic framework for visual localization in ambiguous scenes,'' in \emph{2023 IEEE International Conference on Robotics and Automation (ICRA)}, 2023, pp. 3969--3975.

\bibitem{gramkow2001averaging}
C.~Gramkow, ``On averaging rotations,'' \emph{Journal of Mathematical Imaging and Vision}, vol.~15, no. 1-2, pp. 7--16, 2001.

\bibitem{schonberger2016structure}
J.~L. Schonberger and J.-M. Frahm, ``Structure-from-motion revisited,'' in \emph{IEEE conference on computer vision and pattern recognition}, 2016.

\bibitem{he2016deep}
K.~He, X.~Zhang, S.~Ren, and J.~Sun, ``Deep residual learning for image recognition,'' in \emph{IEEE conference on computer vision and pattern recognition}, 2016, pp. 770--778.

\bibitem{sattler2018benchmarking}
T.~Sattler, W.~Maddern, C.~Toft, A.~Torii, L.~Hammarstrand, E.~Stenborg, D.~Safari, M.~Okutomi, M.~Pollefeys, J.~Sivic \emph{et~al.}, ``Benchmarking 6dof outdoor visual localization in changing conditions,'' in \emph{IEEE conference on computer vision and pattern recognition}, 2018.

\end{thebibliography}
